\documentclass[journal,twoside,web]{ieeecolor}
\usepackage{tmi}
\usepackage{cite}
\usepackage{amsmath,amssymb,amsfonts}
\usepackage{algorithmic}
\usepackage{graphicx}
\usepackage{textcomp}
\usepackage{booktabs}
\usepackage{multirow}
\usepackage[super]{nth}
\usepackage[lined,boxed,ruled,linesnumbered]{algorithm2e}
\usepackage{hyperref}

\newlength\mylen

\newcommand\Tau{\mathcal{T}}% Caligraphic T
\newcommand\Loss{\mathcal{L}}% Caligraphic T 
\def\BibTeX{{\rm B\kern-.05em{\sc i\kern-.025em b}\kern-.08em
    T\kern-.1667em\lower.7ex\hbox{E}\kern-.125emX}}
    
\begin{document}
\title{Self-supervised Learning for Medical Images by Solving Multimodal Jigsaw Puzzles}
\author{Aiham Taleb, Christoph Lippert, Tassilo Klein, and Moin Nabi 
%\IEEEmembership{Fellow, IEEE}
    \thanks{This work is submitted to IEEE Transactions on Medical Imaging - Special Issue on Annotation-Efficient Deep Learning for Medical Imaging, on August the \nth{1}, 2020. }
    \thanks{Aiham~Taleb and Christoph~Lippert are with the Digital Health and Machine Learning Group, Hasso-Plattner Institute, Potsdam University, Rudolf-Breitscheid-Stra{\ss}e 187, 14482 Potsdam, Berlin, Germany.  (e-mail: \texttt{\{aiham.taleb, christoph.lippert\}@hpi.de}). }
    \thanks{Tassilo~Klein and Moin~Nabi are with SAP AI Research, Berlin, Germany
    (e-mail: \texttt{\{tassilo.klein, m.nabi\}@sap.com}). }
}

\maketitle

\begin{abstract}
Self-supervised learning approaches leverage unlabeled samples to acquire generic knowledge about different concepts, hence allowing for annotation-efficient downstream task learning.
In this paper, we propose a novel self-supervised method that leverages multiple imaging modalities. 
We introduce the \emph{multimodal} puzzle task, which facilitates rich representation learning from multiple image modalities. The learned representations allow for subsequent fine-tuning on different downstream tasks.
To achieve that, we learn a modality-agnostic feature embedding by confusing image modalities at the data-level. 
Together with the Sinkhorn operator, with which we formulate the puzzle solving optimization as permutation matrix inference instead of classification, they allow for efficient solving of multimodal puzzles with varying levels of complexity.
In addition, we also propose to utilize cross-modal generation techniques for multimodal data augmentation used for training self-supervised tasks. 
In other words, we exploit synthetic images for \emph{self-supervised pretraining}, instead of downstream tasks directly, in order to circumvent quality issues associated with synthetic images, while improving data-efficiency and the representations learned by self-supervised methods.
Our experimental results, which assess the gains in downstream performance and data-efficiency, show that solving our multimodal puzzles yields better semantic representations, compared to treating each modality independently. Our results also highlight the benefits of exploiting synthetic images for self-supervised pretraining. 
We showcase our approach on four downstream tasks: \emph{Brain tumor segmentation} and \emph{survival days prediction} using four MRI modalities, \emph{Prostate segmentation} using two MRI modalities, and \emph{Liver segmentation} using unregistered CT and MRI modalities. We outperform many previous solutions, and achieve results competitive to state-of-the-art. 
\end{abstract}


\section{Introduction} \label{introduction}
\IEEEPARstart{M}{odern} medical diagnostics heavily rely on the analysis of multiple imaging modalities, particularly, for differential diagnosis~\cite{differential_diagnosis}. However, to leverage the data for supervised machine learning approaches, it requires annotation of large numbers of training examples. Generating expert annotations of patient multimodal data at scale is non-trivial, expensive, time-consuming, and is associated with risks on privacy leakages. Even semi-automatic software tools~\cite{annotate} may fail to sufficiently reduce the annotation expenses.
Consequently, scarcity of data annotations is one of the main impediments for machine learning applications in medical imaging.
At the same time, modern deep learning pipelines are drastically increasing in depth, complexity, and memory requirement, yielding an additional computational bottleneck.

%% A "teaser" image 
\begin{figure}
  \includegraphics[width=\linewidth]{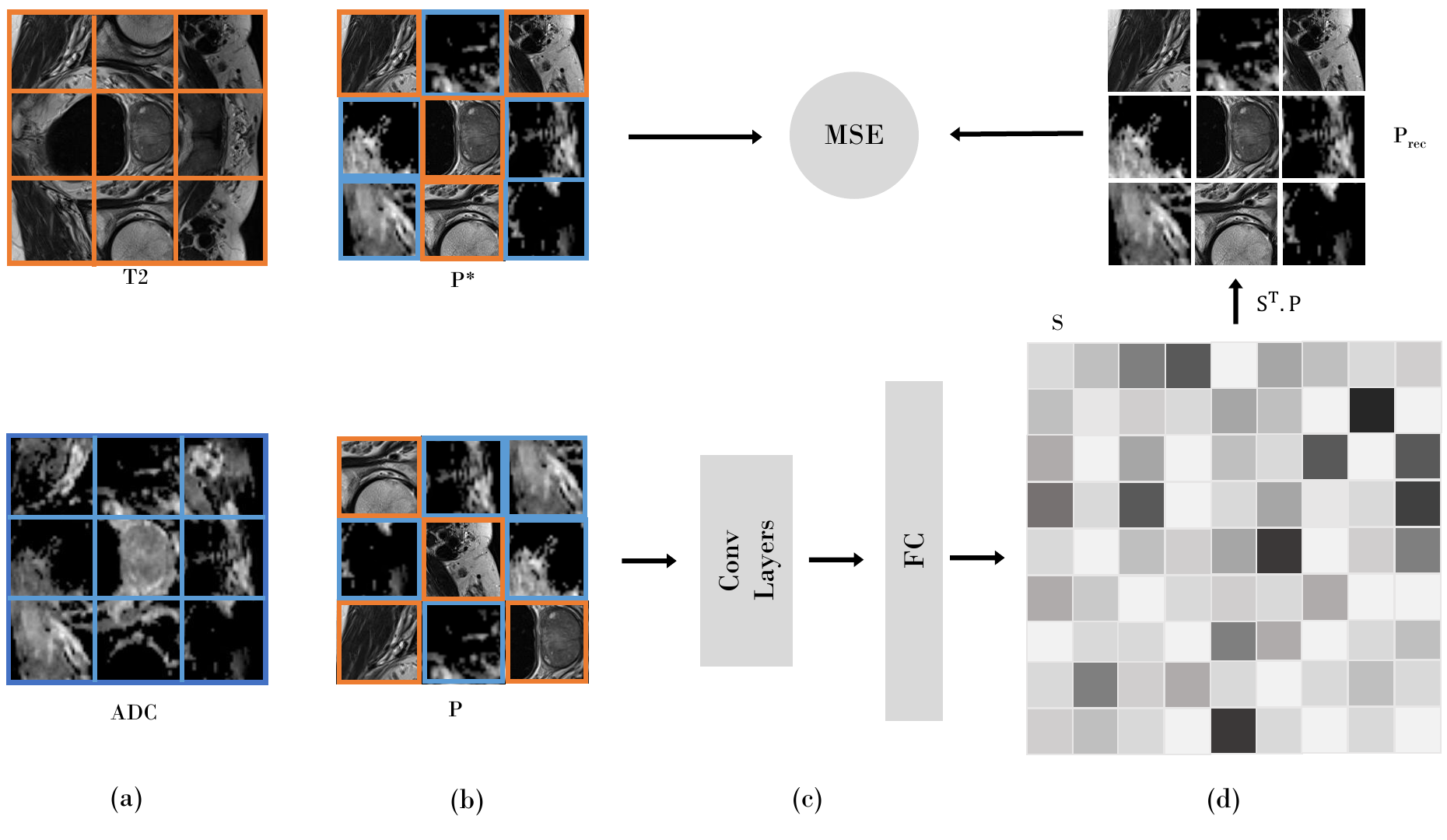}
  \caption{Overview of the process pipeline of the proposed approach, assuming two imaging modalities, e.g., T2 and ADC. (a) multimodal puzzle creation by random selection of each patch from a different random modality, (b) yielding ground truth $P^*$ and random puzzle $P$, (c) randomized puzzles $P$ are injected into the model to train the puzzle-solver with the objective of recovering $P^*$, (d) by applying the learned permutation matrix $S$ to reconstruct $P_{rec}$. The above scans are from the Prostate~\cite{prostate} dataset.}
  \label{fig_small}
\end{figure}

Self-supervised learning provides a viable solution when labeled training data is scarce. In these approaches, the supervisory signals are derived from the data itself, typically by the unsupervised learning of a proxy task. Subsequently, models obtained using self-supervision facilitate data-efficient supervised fine-tuning on the target downstream task, hence reducing the burden of manual annotation. 
Recently proposed self-supervised methods, e.g. \cite{context_prediction,jig}, utilize the spatial context as a supervisory signal to learn effective data representations. However, these approaches do not consider an important characteristic of medical images: their multimodality, e.g. MRI and CT. From an anatomical perspective, multimodality is essential because differences in the physical properties of organs and tissues are translated in a complementary fashion in these multiple modalities. Examples of such cases are numerous~\cite{medical_imaging}: soft body tissues are better encoded in MRI, but CT scans capture bone structures better. Also, specific brain tissues or tumors are better seen in specific MRI modalities. There is a multitude of medical imaging modalities, and in general, the learned representations can be enriched with cross-modal information. Such information is necessary for solving downstream tasks accurately, e.g. semantic segmentation. Thus, we propose to include multiple imaging modalities when designing our self-supervised multimodal Jigsaw puzzle task, to integrate the cross-modal information in the learned representations. The proposed multimodal puzzles are constructed by mixing patches from multiple medical imaging modalities, hence utilizing a modality confusion loss that enables learning powerful cross-modal representations.

Self-supervised pretext tasks, in general, aim to recover the applied transformations to input data. The main intuition behind solving Jigsaw puzzles for visual representation learning is that performing well on this task requires understanding scenes and objects. In other words, a good representation for this task requires learning to identify objects and their parts in order to recover the non-shuffled original images. In our proposed multimodal Jigsaw puzzles, we randomly draw puzzle patches from several imaging modalities, and mix them in the same puzzle. This encourages the model to confuse these imaging modalities, and consequently derive modality-agnostic representations about the data. The interaction between puzzle solving as a task, which aims to learn the underlying structure of the objects, and the modality confusion loss, which aims to learn modality-agnostic representations, encourages the model to derive better notions about the data. By introducing the modality confusion mechanism to Jigsaw puzzle solving, our aim is to account for the inductive bias associated with single-modal puzzles, where objects are believed to have a certain view by the model. Instead, our multimodal puzzles encourage the model derive modality-invariant views about the objects, such as organ structures and tissues. 

While the learned representations by our multimodal puzzles prove useful in several downstream tasks when trained and fine-tuned using \emph{realistic} multimodal images, as shown in our experimental results in Sections~\ref{transfer_learning} and \ref{low_shot}. We also propose to utilize a cross-modal generation step to enhance the quantities of multimodal data samples used in training the puzzle solver. Not only this step allows for better adoption of our proposed approach in real-world scenarios, but also demonstrates the possibility of utilizing \emph{synthetic} images for self-supervised pretraining, instead of directly employing them in downstream task training. To this end, we employ a CycleGAN~\cite{cyclegan} image-to-image translation model, which in addition makes the registration of modalities unnecessary. This step is motivated by clinical scenarios, where data is often non-registered, and the quantities of modalities in the data may vary, i.e. some modalities are more abundant than others, creating a modality imbalance problem. By introducing this step in our pipeline, the imbalance issue is alleviated, as our experimental results in Section~\ref{cross_modal} show.

\textbf{Contributions.} Our contributions are two-fold: \\
\emph{1)} a novel self-supervised multimodal puzzle-solving task, which confuses multiple imaging modalities at the data-level. This allows for combining the complementary information across the modalities about different concepts in the data. In addition, since the modality confusion occurs at data-level, our multimodal puzzles can handle large modality-differences. Our experimental results show that the proposed task can learn rich representations, even when the modalities are non-registered. \\
\emph{2)} we propose to exploit cross-modal image generation (translation) for self-supervised tasks, instead of downstream tasks directly. In order to circumvent the quality issues related to synthetic data, while retaining performance gains in difficult real-world scenarios.
% To the best of our knowledge, this is the first attempt to employ generated data in self-supervised \emph{pretraining}.
Our results show that exploiting inexpensive solutions similar to ours can provide gains in medical image analysis tasks, particularly in low data regimes.

This paper is an extension of our work originally reported in the Medical Imaging Workshop at the Neural Information Processing Systems (NeurIPS 2019) conference~\cite{taleb_ws_paper}. In that preliminary account, our early experiments showed the effectiveness of our proposed multimodal puzzles on brain tumor segmentation using four MRI modalities. In this paper, however, we present the following extensions:
\begin{enumerate}
    \item We propose to employ cross-modal generation as an additional step to address address real-world clinical situations where multimodal data is not available in abundance, i.e. modality imbalance is an issue. Yet, learning cross-modal representations allows solving downstream tasks more accurately.
    % As mentioned earlier, in our setting, we use synthetic data for self-supervised pretraining, aiming to improve the learned representations, while circumventing quality issues associated with synthetic data when used in downstream tasks directly. 
    \item A comprehensive evaluation of the proposed multimodal puzzles on additional datasets for the downstream tasks of: i) Prostate segmentation using two MRI modalities, and ii) Liver segmentation using CT and MRI modalities. Both tasks highlight the ability of our proposed puzzles to learn powerful representations when the difference between the modalities is high, due to confusing the modalities at the data-level. The Liver segmentation task, in particular, demonstrates our task's ability to learn rich representations when the modalities are non-registered (unaligned or unpaired). 
\end{enumerate}

%------------------------------------------------------------------------
\section{Related Work} \label{related_work}
\textbf{Self-supervised learning} methods construct a representation (embedding) space by creating a supervised proxy task from the data itself. The embeddings that solve the proxy task will also be useful for other real-world downstream tasks, afterwards. These methods differ in their core building block, i.e. the proxy task used to learn representations from unlabelled input data. A commonly used supervision source for proxy tasks is the spatial context from images, which was first inspired by the skip-gram Word2Vec~\cite{w2v} algorithm. This idea was generalized to images in the work of Doersch \emph{et al.}~\cite{context_prediction}, in which a visual representation is learned by the task of predicting the position of an image patch relative to another. Examples of works that utilize the spatial context are numerous~\cite{jig,ssl_360_images}. Other self supervised methods used different supervision sources, such as image colors~\cite{color}, clustering~\cite{deep_cluster}, image rotation prediction~\cite{rotations}, object saliency~\cite{ssl_salient_beauty}, and image reconstruction~\cite{ssl_models_genesis}. In more recent works, Contrastive Learning approaches~\cite{CPC1,CPC2,chen2020simple} advanced the results of self-supervised methods on the ImageNet benchmark~\cite{imagenet_cvpr09}. Our work follows this line of research, and it utilizes the spatial context as a source of supervision. However, this context is derived across multiple imaging modalities, encouraging the model to learn modality-agnostic notions about the data. 

The closest self-supervised method to ours is that of Noroozi \emph{et al.}~\cite{jig}, which extended the patch-based approach of~\cite{context_prediction} to solve Jigsaw puzzles on natural images as a proxy task. In contrast to our approach, this method only relies on a single image modality, which limits its ability to capture the vital cross-modal complementary information.
In addition, this method requires massive memory and compute resources, even for small puzzles of 3-by-3 as it uses 9 replicas of AlexNet~\cite{alexnet}. 
In our method, on the other hand, we utilize multiple imaging modalities in our puzzle solving task. Also, to improve the computational tractability, our approach utilizes the efficient Sinkhorn operator~\cite{adams,sinkhorn_op,mena2018learning} as an analog to the Softmax, but for permutation-related tasks. Consequently, our approach can solve puzzles with more levels of complexity. The Sinkhorn operator allows for casting the puzzle solving problem as a permutation matrix inference instead of classification, which relies on choosing a suitable permutation set as classification targets.
In Noroozi \emph{et al.}'s method~\cite{jig}, the choice of a fixed permutation set size defines the complexity of the classification task. However, this fixed set size limits the complexity of the self-supervised task. On the other hand, by defining our task as a permutation matrix inference, the model is enforced to find the applied permutation among \emph{all} possible permutations. 

\textbf{Solving Jigsaw puzzles} is a common pattern recognition task in computer vision. Several puzzle solving algorithms have been proposed in the area of artificial intelligence for different purposes and applications. Greedy algorithms~\cite{puzzle_solver_greedy_1,puzzle_solver_greedy_2,puzzle_solver_greedy_3} rely on sequential pairwise patch matching to solve Jigsaw puzzles, which can be computationally inefficient. More recent approaches search for global solutions that observe all the patches, and minimize a measure over them. Example methods are numerous: such as Probabilistic methods using Markov Random Fields~\cite{puzzle_solver_probabilistic}, Genetic algorithms~\cite{puzzle_solver_genetic}, algorithms for consensus agreement across neighbors~\cite{puzzle_solver_consensus}, and solving it as a classification problem~\cite{jig}. The latter work, from Noroozi \emph{et al.}~\cite{jig}, allows for data-efficient representation learning. However, as mentioned earlier, such puzzle solver is capable to capture only a subset of possible permutations of the image tiles. Hence, solutions that capture the whole set of permutations~\cite{adams,mena2018learning,puzzle_solver_perm_approx1,puzzle_solver_perm_approx2} are a better alternative. These solutions, in general, approximate the permutation matrix, which is a doubly stochastic matrix, and solve an optimization problem to recover the right matrix. Each permutation, e.g. of puzzle tiles, corresponds to a unique permutation matrix, which is the ground truth of such models and is approximated by differentiable operators similar to Sinkhorn~\cite{sinkhorn_op}, as detailed later. The main advantage of such operators is that they can be formulated, similar to the Softmax function, in the output of a neural network, allowing for efficient and cheap solving for Jigsaw puzzles~\cite{adams}. Puzzle solving has been cast as a pixel-wise image reconstruction problem in~\cite{puzzle_solver_unet} using a U-Net architecture and a VAE in~\cite{puzzle_solver_vae}, and a GAN in~\cite{puzzle_solver_gan}. We believe pixel-wise image reconstruction/generation tasks may influence the representations with low-level stochastic texture details, rather than high-level feature representations. In addition, pixel-level generation is computationally expensive and may not be necessary for representation learning. We compare to a sample approach to validate this behavior.

Applications of solving jigsaw puzzles as a proxy task include domain adaptation across different datasets~\cite{domain_adaptation_caputo,domain_adaptation_efros}. In a multi-task learning scheme, both works exploit jigsaw puzzles as a secondary task for object recognition, acting mainly as a regularizer. Their aim is to learn \emph{domain}-agnostic representations for different concepts, we utilize jigsaw puzzle solving in an attempt to learn \emph{modality}-agnostic representations, however. Also, as opposed to our approach, both works use self-supervision in a multi-task fashion, solving jointly the same tasks on multiple domains. In this scenario, the model is expected to confuse the modalities/domains at the \emph{feature-level}, i.e. in the learned features. This type of domain/modality interaction is similar to the late fusion of multimodal data~\cite{multimodal_survey}. On the other hand, we create a multimodal task by fusing the data of multiple modalities and then solving that task, i.e. we perform an early modality fusion~\cite{multimodal_survey}.
Their approach is likely to fail when the domain difference is high, i.e. the modality difference in our case. On the other hand, our approach can handle this issue as the integration of modalities occurs at the \emph{data-level}, as shown in our experimental results. 

\textbf{Multimodal learning} works on are numerous~\cite{multimodal_survey,multimodal_dl}, and attempt to tackle many of its inherent difficulties, such as: multimodal fusion, alignment, and more importantly, representation. Multimodal representations allow for solving several tasks, which are otherwise unfeasible using single-modal representations. Prior works, some of which are self-supervised methods, addressed tasks that involve various combinations of modalities, such as: image and text~\cite{vqa,vqa_medical,image_caption1,image_caption2,lip_reading,text2image,cross_modal_scenes}, image and audio~\cite{audio_animation,audio_image1,audio_image2,audio_image3,soundnet,emotion_cross_modal,speech_grounding}, audio and text~\cite{audio_text1,audio_text2}, and multiview (multimodal) images~\cite{Multimodal_decision_making,cross_learn,rgb_optical1,rgb_optical2}. The latter works are the most similar to ours in terms of modality types they combine. However, our approach fuses the imaging modalities at the \emph{data-level}, as opposed to the \emph{feature-level} used in these works. As explained above, our approach can handle large modality differences, and thus enables better modality confusion. 

\textbf{Self-supervision in the medical context.} Self-supervision has found use cases in diverse applications such as depth estimation in monocular endoscopy~\cite{depth}, medical image registration~\cite{register}, body part recognition~\cite{body}, in disc degeneration using spinal MRIs~\cite{disc}, and body part regression for slice ordering~\cite{Yan2018DeepLG}. Many of these works make assumptions about input data, resulting in engineered solutions that hardly generalize to other target tasks. Our proposed approach avoids making such assumptions about input imaging modalities. Instead, our experimental results show that the proposed multimodal puzzles may operate on different imaging modalities, even when they are spatially unregistered.
In a more related set of works, Tajbakhsh \emph{et al.}~\cite{orientation_prediction_tajbakhsh} use orientation prediction from medical images as a proxy task, Spitzer \emph{et al.}~\cite{spitzer_3d_distance} predict the 3D distance between two 2D brain patches, Zhou \emph{et al.} use image reconstruction (restoration) to derive the supervision signals, Ahn \emph{et al.}~\cite{ssl_domain_adaptation} and Pan \emph{et al.} \cite{ssl_feature_augmentation} use feature augmentation on medical images, Jiao \emph{et al.}~\cite{ultrasound_video} use the order of ultrasound video clips and random geometric transformations, and Ahn \emph{et al.}~\cite{ssl_kernel_learning} utilize sparse kernel-learning and a layer-wise pretraining algorithm. Zhou \emph{et al.}~\cite{ssl_models_genesis} extended image reconstruction techniques from 2D to 3D, and implemented multiple self-supervised tasks based on image-reconstruction. Zhou \emph{et al.}~\cite{ssl_models_genesis} extended image reconstruction techniques from 2D to 3D, and implemented multiple self-supervised tasks based on image-reconstruction. Taleb \emph{et al.}~\cite{taleb_3D_paper} extended five self-supervised methods to the 3D context, and showed gains in the medical imaging domains.
A review of similar works is provided in~\cite{review_annotation_efficient}. 
We follow this line of research, and we propose to exploit multiple imaging modalities in designing proxy tasks. Zhuang \emph{et al.}~\cite{rubik} developed a proxy task for solving 3D jigsaw puzzles, in an attempt to simulate Rubik's cube solving. Since their work is an extension of the 2D puzzles of Noroozi \emph{et al.}~\cite{jig} to 3D, it incurs similar computational costs, only the issue here is exacerbated as 3D puzzles require more computations. This explains their limitation to only $2\times2\times2$ puzzles. Our multimodal puzzle task outperforms their method, while requiring less computations.

\textbf{Image-to-image translation} using Generative Adversarial Networks (GANs)~\cite{pix2pix,cyclegan,DualGAN} has found several use cases in the medical domain. Many works have attempted to translate across medical imaging modalities~\cite{mr_to_ct,medgan,mr_to_ct2,mr_syn,cyclegan3d}. While the goal of these methods is to improve the cross-modal generation quality, we view it as orthogonal to our goal. Similar to~\cite{sandfort_cyclegan,fu_cyclegan,tang_cyclegan}, we utilize cross-modal translation methods to improve the performance on downstream tasks, e.g. segmentation, through augmentation. However, especially for clinical applications, one can doubt the quality of synthetic images. Hence, as opposed to the above works, we circumvent this issue by employing these images for pretraining purposes only, and not the final prediction layer. Furthermore, this fashion of exploiting generated images allows for solving cases where multimodal samples are a few, yet required to obtain cross-modal information. For instance, in a clinical setting, it is common to find specific modalities in abundance (e.g. X-Ray or CT), and some in smaller quantities (such as MRI), e.g. due to acquisition costs. Cross-modal translation techniques have the potential to address such scenarios. 

%------------------------------------------------------------------------
\section{Method} \label{method}
Our method processes input samples from datasets that contain multiple imaging modalities, as it is the case in the majority of medical imaging datasets. The types of medical imaging modalities are numerous~\cite{medical_imaging}, and they vary in their characteristics and use-cases. We assume no prior knowledge about what modalities are being used in our models, i.e. the modalities can vary from one downstream task to another.
\begin{figure*}[htb]
\begin{center}
    \includegraphics[width=0.9\linewidth]{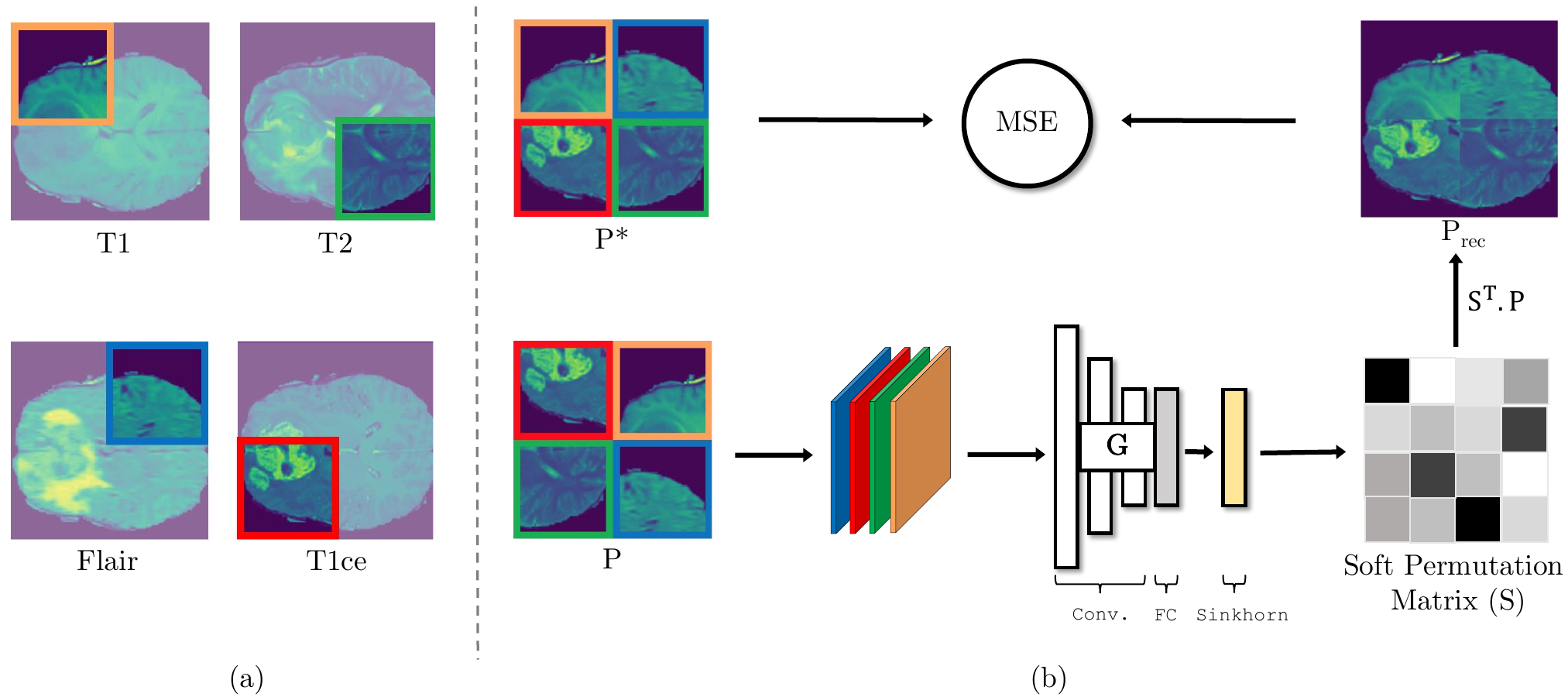}
\end{center}
\caption{Schematic illustration showing the steps of the proposed multimodal puzzles. (a) Assuming we have four modalities here, even though our puzzles can be created from any number of image modalities depending on the downstream task, (b) these images are then used to construct multimodal jigsaw puzzles, drawing patches from all the modalities randomly. These scans are from the BraTS~\cite{brats1,brats2} dataset.} \label{fig_big}
\end{figure*}

\subsection{Multimodal Puzzle Construction} \label{puzzle_construction}
Solving a jigsaw puzzle entails two main steps. First, the image is cut into puzzle pieces (patches or tiles), which are shuffled randomly according to a certain permutation. Second, these shuffled image pieces are assembled such that the original image is restored. If $C$ is the number of puzzle pieces, then there exist $C!$ of possible puzzle piece arrangements. It should be noted that when the puzzle complexity increases, the association of individual puzzle tiles can be ambiguous, e.g. puzzle tiles that originate from uni-colored backgrounds can be tricky to place correctly.
Nevertheless, the placement of different puzzle tiles is mutually exclusive. Thus, when all tiles are observed at the same time, the positional ambiguities are alleviated.
In a conventional jigsaw puzzle, the puzzle pieces originate from only one image at a time, i.e. the computational complexity for solving such a puzzle is $O(C!)$.

On the other hand, we propose a \emph{multimodal} jigsaw puzzle, where tiles can be from $M$ different modalities. This proposed multimodal puzzle simultaneously learns the in-depth representation of how the organs compose, along with the spatial relationship across modalities.
As a result, the complexity of solving multimodal puzzles is increased to $O(C!^M)$. Consequently, this quickly becomes prohibitively expensive due to two growth factors in the solution space: i) factorial growth in the number of permutations $C!$, ii) exponential growth in the number of modalities $M$. To reduce the computational burden, we use two tricks. First, we employ the Sinkhorn operator, which allows for an efficient solving of the factorial factor. Second, we employ a feed-forward network $G$ that learns a cross-modal representation, which allows for canceling out the exponential factor $M$ while simultaneously learning a rich representation for downstream tasks.

% \begin{algorithm} 
%     \SetKwProg{Algorithm}{Algorithm}{}{}
%     \SetKwProg{Procedure}{Procedure}{}{}
%     \SetNlSty{textbf}{}{:}
%     \DontPrintSemicolon
%     \Algorithm{\textsc{Create Puzzles}}{
%       \KwIn{- modality lists ($m_1, m_2, \dots, m_M$), each with $L$ slices}
%       \myinput{- number of patches in a puzzle ($np$)}
%       \myinput{- list of possible permutations ($perms$)}
%       \myinput{- \# of puzzles to generate per slice ($nps$)}
%       \KwOut{list of multimodal $puzzles$}
%       \For{$i\leftarrow 1$ \KwTo $L$}{
%         \For{$pt\leftarrow 1$ \KwTo $np$}{
%           $m \leftarrow$ choose random modality\;
%           $patches[pt] \leftarrow$ fill patch in position $pt$ from slice with modality $m$\;
%         }
%         \For{$p\leftarrow 1$ \KwTo $nps$}{
%           $perm\_patches \leftarrow$ shuffle $patches$ using a random permutation from $perms$\;
%           append $perm\_patches$ to $puzzles$\;
%         }
%       }
%       \KwRet{$puzzles$}\;
%     }
%     \caption{Multimodal jigsaw puzzle creation} \label{algo1}
% \end{algorithm}

\subsection{Puzzle-Solving with Sinkhorn Networks}
To efficiently solve the self-supervised jigsaw puzzle task, we train a network that can learn a permutation. A permutation matrix of size $N \times N$ corresponds to some permutation of the numbers $1$ to $N$. Every row and column, therefore, contains precisely a single $1$ with $0$s everywhere else, and every permutation corresponds to a unique permutation matrix. This permutation matrix is non-differentiable. However, as shown in \cite{mena2018learning}, the non-differentiable parameterization of a permutation can be approximated in terms of a differentiable relaxation, the so-called Sinkhorn operator. 
The Sinkhorn operator iteratively normalizes rows and columns of any real-valued matrix to obtain a “soft” permutation matrix, which is doubly stochastic. Formally, for an arbitrary input $X$, which is an $N$ dimensional square matrix, the Sinkhorn operator $S(X)$ is defined as:
\begin{equation}
\begin{split}
S^{0}(X) &= exp(X), \\
S^{i}(X) &= \Tau_{R}(\Tau_{C} (S^{i-1}(X))), \\
S(X) &= \lim_{i \rightarrow\infty} S^{i}(X).
\end{split}
\end{equation}
where $\Tau_{R}(X) = X \oslash (X\mathbf{1}_{N}\mathbf{1}^\top_N)$ and $\Tau_{C}(X) = X \oslash (\mathbf{1}_{N}\mathbf{1}^\top_N X)$ are the row and column normalization operators, respectively. The element-wise division is denoted by $\oslash$, and $\mathbf{1}^\top_N \in \mathbb{N}^N$ is an $N$ dimensional vector of ones.

Assuming an input set of patches $P = \{p_1, p_2, ..., p_N\}$, where $P \in \mathbb{R}^{N \times l \times l}$ represents a puzzle that consists of $N$ square patches, and $l$ is the patch length. We pass each element in $P$ through a network $G$, which processes every patch independently and produces a single output feature vector with length $N$. By concatenating together these feature vectors obtained for all region sets, we obtain an $N \times N$ matrix, which is then passed to the Sinkhorn operator to obtain the soft permutation matrix $S$. Formally, the network $G$ learns the mapping $G: P \rightarrow S$, where $S \in [0,1]^{N \times N}$ is the soft permutation matrix, which is applied to the scrambled input $P$ to reconstruct the image $P_{rec}$. The network $G$ is then trained by minimizing the mean squared error (MSE) between the sorted ground-truth $P^*$ and the reconstructed version $P_{rec}$ of the scrambled input, as in the loss formula below: 
\begin{equation}
\Loss_{puzzle} (\theta , P, P^*) = \sum_{i=1}^{K}\left \| P_i^* - S^T_{\theta,P_i} . P_i \right \|^{2},
\end{equation}
where $\theta$ corresponds to the network parameters, and $K$ is the total number of training puzzles. After obtaining the network parameters $\theta$, the yielded representations capture different tissue structures across the given modalities as a consequence of the multimodal puzzle solving. Therefore, the learned representations can be employed in downstream tasks by a simple fine-tuning on target domains, in an annotation-efficient regime. Our proposed approach is depicted in figure~\ref{fig_big}.

\subsection{Cross-Modal Generation}
\begin{figure}
  \includegraphics[width=\linewidth]{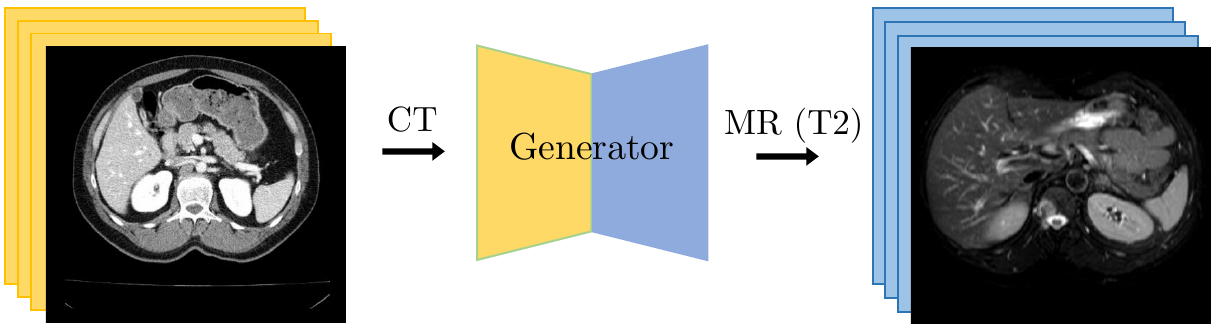}
  \caption{A cross-modal generation example on abdominal scans from CT to MR (T2) modalities. Here, $G$ is a generator model between these modalities. These scans are from the CHAOS~\cite{liver} dataset.}
  \label{fig_small_gan}
\end{figure}

Multimodal medical images exist in several curated datasets, and in pairs of aligned (or registered) scans. However, as described before, in many real-world scenarios, obtaining such data in large quantities can be challenging. To address this, we add an explicit cross-modal generation step using CycleGAN~\cite{cyclegan}, illustrated in figure~\ref{fig_small_gan}. This model also uses a cycle-consistency loss, which allows for relaxing the alignment (pairing) constraint across the two modalities. Therefore, CycleGAN can translate between any two imaging modalities, requiring no prior expensive registration steps. This step allows us to leverage the richness of multimodal representations obtained by our proposed puzzle-solving task. In our scenario, after generating data samples of the small (in number of samples) modality using samples from the larger modality, we construct our multimodal puzzles using a mix of real and generated multimodal data.  As we show in our experiments, this yields better representations compared to using a single modality only when creating puzzles. We have to highlight that, our multimodal puzzles are capable of operating on real multimodal images merely, which are the results shown in Section~\ref{transfer_learning}. However, we assess the influence of mixing those real multimodal images with synthetic ones in Section~\ref{cross_modal}.

%------------------------------------------------------------------------
\section{Experimental Results} \label{experiments}
In the following sections, we assess the performance of our proposed pretraining method on multimodal medical imaging datasets detailed in Section~\ref{datasets}. We transfer (and fine-tune) the learned representations by our model to different downstream tasks, and measure their impact in Section~\ref{transfer_learning}. Then, we study the effect of integrating generated data in constructing our multimodal puzzles in Section~\ref{cross_modal}. Next, we assess how our self-supervised task affects the downstream tasks' data efficiency, i.e. when operating in a low-data regime, in Section~\ref{low_shot}. Finally, we analyze the effect of many variables (puzzle complexity and permutation list size) in our multimodal puzzles on downstream task performance in an ablation study in Section~\ref{puzzle_complexity}.  

\subsection{Datasets} \label{datasets}
In our experiments, we consider three multimodal medical imaging datasets. The first is the Multimodal Brain Tumor Image Segmentation Benchmark (\textbf{BraTS}) dataset~\cite{brats1,brats2}. This dataset is widely used to benchmark different semantic segmentation algorithms in the medical imaging domain. It contains multimodal MRI scans for 285 training cases and for 66 validation cases. All BraTS scans include four MRI modalities per case: a) native (T1), b) post-contrast T1-weighted (T1Gd), c) T2-weighted (T2), and d) T2 Fluid Attenuated Inversion Recovery (T2-FLAIR) volumes. The BraTS challenge involves two different tasks: i) brain tumor segmentation, and ii) number of survival days prediction. 

The second benchmark is the \textbf{Prostate} segmentation task from the Medical Segmentation Decathlon~\cite{prostate}. The prostate dataset consists of 48 multimodal MRI cases, from which 32 cases are used for training, and 16 are used for testing. Manual segmentation of the whole prostate was produced from T2-weighted scans, and the apparent diffusion coefficient (ADC) maps. The target challenge is for segmenting two adjoint prostate regions (the central gland and the peripheral zone).

The third benchmark is the \textbf{Liver} segmentation task from the CHAOS~\cite{liver} multimodal dataset. The CHAOS dataset consists of 40 multimodal cases, from which 20 cases are used for training, and 20 for testing. This dataset consists of CT and MRI multimodal data, where each case (patient) has a CT and an MRI scans. The main task in this benchmark is liver segmentation, and manual segmentation masks exist for each case. We only benchmark our models on the liver segmentation task, to demonstrate our approach on this main task. Even though other organs in the abdomen are annotated in MRI scans, but not in CT. Hence, segmenting the other organs, aside from the liver, is beyond the scope of this work. The CT and MR modalities in this benchmark are not only different in appearance, but also they are non-registered, making this dataset a pertinent test-bed for our multimodal puzzles.

\subsection{Transfer Learning Results} \label{transfer_learning}
We evaluate the quality of the learned representations from our task of multimodal puzzle-solving by transferring them in downstream tasks. Then, we assess their impact on downstream performance. As mentioned before, we only use \emph{realistic} data in the experiments presented in this section. 

\subsubsection{Brain Tumor Segmentation} 
The goal of this task is to segment 3 different regions of brain tumor: a) the whole tumor (WT), b) the tumor core (TC), and c) the enhanced tumor (ET). Each of these regions has different characteristics, and each may appear more clearly on specific MRI modalities than others, justifying the need for multiple modalities.

\paragraph{Baselines}
In order to assess the quality of our representations, we establish the following baselines. Apart from the \textbf{Single-modal} baseline, all of the following baselines use multimodal data. To average out random variations and compute \textit{p}-values, we used a 5-fold cross validation evaluation approach. For each fold, we used the same training and test datasets for our method and all baselines to perform a one-sided two-sample t-test on the dice scores, following~\cite{Multimodal_decision_making}.

\textbf{From Scratch:} The first sensible baseline for all self-supervised methods is to compare with the model when trained on the downstream task from scratch. This baseline provides an insight into the benefits of self-supervised pretraining (initialization), as opposed to learning the target task directly.

\textbf{Single-modal:} We study the impact of our pretraining method on this task when processing only a single modality as input. This experiment aims at simulating the realistic situation when human experts examine brain scans, as some modalities highlight certain aspects of the tumor more than others. For instance, Flair is typically used to examine the whole tumor area, while T2 is used for tumor core, and the T1ce highlights the enhanced tumor region. We select the best modality for each task when comparing to these results. 

\textbf{Isensee \emph{et al.}~\cite{isensee}:} This work ranks among the tops in the BraTS 2018 challenge. It uses additional datasets next to the challenge data, and it performs multiple types of augmentation techniques. The model architecture is a 3D U-Net~\cite{UNET}. We only fine-tune our learned representations from the self-supervised task, thus requiring much less data and augmentation methods. \textbf{2D Isensee} is a 2D version of their network, which we implement for better comparability.

\textbf{Chang~\emph{et al.}~\cite{chang}:} Trained multiple versions of the 2D U-Net models, and used them as an ensemble to predict segmentation masks. This requires significantly more computing time and resources than training a single model that performs the task with higher performance in many cases. 

\textbf{Li~\cite{Li2018}:} Implemented a 3-stage cascaded segmentation network that combines whole-tumor, tumor-core and enhanced-tumor masks. For the whole-tumor stage, they utilize a modified multi-view 2D U-Net architecture, which processes three slices at a time from input 3D scans: axial, sagittal, and coronal views. Our method produces better results while requiring less computations using a smaller network.

\textbf{JiGen~\cite{domain_adaptation_caputo}:} This method is a multi-tasking approach called JiGen~\cite{domain_adaptation_caputo}. JiGen solves jigsaw puzzles as a secondary task for domain generalization. We implemented their model and treated the multiple modalities as if they were other domains. This baseline aims to analyze the benefits of performing modality confusion on the data-level (our approach), as opposed to the feature-level (their approach).

\textbf{Models Genesis~\cite{ssl_models_genesis}:} This is a self-supervised method that relies on multiple image reconstruction tasks to learn from unlabeled scans. Even though their method is mainly implemented in 3D, we employ the 2D version (\textbf{2D MG}) of their model\footnote{It uses Resnet18 as the encoder of the U-Net architecture implemented here: \url{https://github.com/qubvel/segmentation\_models}} (pretrained on Chest-CT), for better comparability. 

\textbf{Rubik Cube~\cite{rubik}:} A self-supervised method that relies on solving 3D jigsaw puzzles as a proxy task, and also applies random rotations on puzzle cubes. Similarly, we compare to a 2D (\textbf{2D RC}) version of their method, for better comparability. 

\textbf{U-Net Puzzle Solver:} This baseline aims to evaluate the representations learned by our puzzle solver, which relies on Sinkhorn operator, in comparison to a U-Net puzzle solver, which relies on pixel-wise image reconstruction. This baseline is trained with a simple pixel-wise MSE loss~\cite{puzzle_solver_unet}.

\paragraph{Evaluation Metrics}
The reported metrics are the average dice scores for the Whole Tumor (WT), the Tumor Core (TC), and the Enhanced Tumor (ET). 

\paragraph{Discussion}
The results of our \textbf{multi-modal} method compared to the above baselines are shown in table~\ref{tab_brain}. Our proposed method outperforms both the "from scratch" and "single-modal" baselines, confirming the benefits of pretraining using our multimodal approach. 
In addition, our method achieves comparable results to methods from literature. We outperform these baselines in most cases, such as the methods Chang \emph{et al.}~\cite{chang}, and Li~\cite{Li2018}, in terms of all reported dice scores. 
We also report the result of a 2D version (for better comparability) of Isensee \emph{et al.}~\cite{isensee}, which ranks among the best results on the BraTS 2018 benchmark. Even though their method uses co-training with additional datasets and several augmentation techniques, we outperform their results in this task. This supports the performance benefits of initializing CNNs with our multimodal puzzles. We also compare with 2D Models Genesis (2D MG)~\cite{ssl_models_genesis}, which we outperform in this downstream task, supporting the effectiveness of our pretraining method. Compared to 2D Rubik Cube~\cite{rubik}, which we implement in 2D for comparability, we observe that we outperform this method in this task. This confirms the higher quality of the representations learned by our method, compared to this jigsaw puzzle solving method (including random rotations). 
Compared to the work of~\cite{domain_adaptation_caputo} (JiGen), we also find that our results outperform this baseline, confirming that our approach of performing the modality confusion in the data-level is superior to modality confusion in the feature-level. 
Our Sinkhorn-based puzzle solver produces representations that are superior to those obtained by a U-Net puzzle solver, as shown in the table. We believe that pixel-wise image reconstruction adds unnecessary complexity to the puzzle solving task, which can be solved simply with an encoder-based CNN architecture, as explained before. 
\begin{table}
\begin{center}
\centering         \small
\caption{Results on the BraTS segmentation task}\label{tab_brain}
\begin{tabular}[t]{ c c c c c } \toprule
Model &  ET & WT & TC & \textit{p}-value\\
\midrule
From scratch & 68.12 & 80.54 & 77.29 &  3.9$e$-5\\
Single-modal & 78.26 & 87.01 & 82.52 & 6.0$e$-4 \\
\hline 
Li~\cite{Li2018} & 73.65 & 88.24 & 78.56 & 9.0$e$-4 \\
Chang \emph{et al.}~\cite{chang} & 75.90 & 89.05 & 82.24 & 2.6$e$-3 \\
2D MG~\cite{ssl_models_genesis} & 79.21 & 88.82 & 83.60 & 7.6$e$-2 \\
2D Isensee \emph{et al.}~\cite{isensee} & 78.92 & 88.42 & 83.73 & 4.6$e$-2 \\
2D RC~\cite{rubik} & 78.38 & 87.16 & 82.92 & 1.4$e$-2 \\
JiGen~\cite{domain_adaptation_caputo} & 78.75 & 88.15 & 83.32 & 5.0$e$-4 \\
U-Net Solver & 70.65 & 82.54 & 75.18 & 1.2$e$-11 \\
\hline 
Ours (Multi-modal) & \textbf{79.65} & \textbf{89.74} & \textbf{84.48} & - \\
\bottomrule
\end{tabular}
\end{center}
\end{table}

%-------------------------------------------------------------------------
\subsubsection{Prostate Segmentation}
The target of this task is to segment two regions of the prostate: central gland, and peripheral zone. This task utilizes two available MRI modalities. 
    
\paragraph{Baselines}
We establish the following baselines, which, apart from \textbf{Single-modal}, all use multimodal data. To average out random variations and compute \textit{p}-values, we used a 5-fold cross validation evaluation approach. For each fold, we used the same training and test datasets for our method and all baselines to perform a one-sided two-sample t-test on the dice scores, following~\cite{Multimodal_decision_making}.

\textbf{From Scratch:} Similar to the first downstream task, we compare our model with the same architecture when training on the prostate segmentation task from scratch. 

\textbf{Single-modal:} We also study the impact of our method when using only a single modality (T2) to create the puzzles.

\textbf{JiGen~\cite{domain_adaptation_caputo}:} Similar to the first downstream task, we compare our method to the multi-tasking approach JiGen.

\textbf{2D Models Genesis~\cite{ssl_models_genesis} (2D MG):} Similar to the first task, we fine-tune this model on multimodal Prostate data.

\textbf{2D Rubik Cube~\cite{rubik} (2D RC):} Similar to the first task, we fine-tune the 2D version of this method on multimodal prostate data.

\textbf{U-Net Puzzle Solver:} This baseline aims to evaluate the representations learned by our puzzle solver, which relies on Sinkhorn operator, in comparison to a U-Net puzzle solver, which relies on pixel-wise image reconstruction.

\paragraph{Evaluation Metrics}
We report the values of 2 evaluation metrics in this task, the average dice score (Dice) and the normalized surface distance (NSD). These metrics are used on the official challenge. The metrics are computed for the 2 prostate regions (\textbf{C}entral and \textbf{P}eripheral).

\paragraph{Discussion}
The results of our \textbf{multi-modal} method compared to the above baselines are shown in table~\ref{tab_prostate}. Our proposed method outperforms both "from scratch" and "single-modal" baselines in this task, too, supporting the advantages of pretraining the segmentation model using our multimodal approach. We also compare with 2D Models Genesis (2D MG)~\cite{ssl_models_genesis}, which we outperform in this downstream task, supporting the effectiveness of our pretraining method. Also, our method outperforms the multitasking method JiGen~\cite{domain_adaptation_caputo}, when trained on this task too. We observe a more significant gap in performance between our approach and JiGen in this task, compared to the first downstream task of brain tumor segmentation. We posit that this can be attributed to the more significant difference between the imaging modalities used in this prostate segmentation task, as opposed to those in the brain tumor segmentation task. The figure~\ref{fig_ql} shows this difference more clearly. It can be noted that the imaging modalities of the prostate dataset, are more different in appearance than those of the brain tumor dataset. This difference in appearance among the modalities can be explained by understanding the physics from which these MRI modalities are created. All of the brain MRI sequences in the BraTS dataset are variants of T1- and T2-weighted scans, they only differ in configurations of the MRI scanner. These different configurations cause the contrast and brightness of some brain areas to vary among these MRI sequences. The ADC map, on the other hand, is a measure of the magnitude of diffusion (of water molecules) within the organ tissue. This requires a specific type of MRI imaging called Diffusion Weighted Imaging (DWI). In general, highly cellular tissues or cellular swellings exhibit lower diffusion coefficients, e.g. a tumor, a stroke, or in our case, the prostate. 
Similar to the first downstream task, our Sinkhorn-based puzzle solver outperforms the U-Net puzzle solver, confirming that the representations obtained by our solver hold more semantic information. Compared to 2D Rubik Cube~\cite{rubik}, similarly, we outperform this method on this downstream task.
 
\begin{table}
\begin{center}
\centering         \small
\caption{Results on the Prostate segmentation task}\label{tab_prostate}
\begin{tabular}[t]{ c c c c c c } \toprule
\multirow{2}{*}{Model} & \multicolumn{2}{c}{Dice} & \multicolumn{2}{c}{NSD} & \multirow{2}{*}{\textit{p}-value} \\
\cline{2-5}
 & C & P & C & P \\
\midrule 
From scratch & 68.98 & 86.15 & 94.57 & 97.84 & 4.9$e$-3 \\
Single-modal & 69.48 & 87.42 & 92.97 & 97.21 & 9.3$e$-5 \\
\hline 
2D MG~\cite{ssl_models_genesis} & 73.85 & 87.77 & 94.61 & 98.59 &  4.3$e$-2 \\
2D RC~\cite{rubik} & 73.11 & 86.14 & 93.65 & 97.47 & 3.1$e$-2 \\
JiGen~\cite{domain_adaptation_caputo} & 69.98 & 86.82 & 92.67 & 96.13 & 9.1$e$-3 \\
U-Net Solver & 64.72 & 84.26 & 88.47 & 90.36 & 9.0$e$-7 \\
\hline 
Ours (Multi-modal) & \textbf{75.11} & \textbf{88.79} & \textbf{94.95} & \textbf{98.65} & - \\
\bottomrule
\end{tabular}
\end{center}
\end{table}

%-------------------------------------------------------------------------
\subsubsection{Liver Segmentation}
The target of this task is to segment the liver from multimodal abdominal scans, which include CT and MRI modalities. 

\paragraph{Baselines}
Apart from the \textbf{Single-modal} baseline, all of the following baselines use multimodal data. To average out random variations and compute \textit{p}-values, we used a 5-fold cross validation evaluation approach. For each fold, we used the same training and test datasets for our method and all baselines to perform a one-sided two-sample t-test on the dice scores, following~\cite{Multimodal_decision_making}.

\textbf{From Scratch:} Similar to the first downstream task, we compare our model with the same architecture when training on liver segmentation from scratch. 

\textbf{Single-modal:} We also study the impact of our pretraining method when using only a single modality to create the puzzles. We choose CT, as it exists in larger quantities.

\textbf{JiGen~\cite{domain_adaptation_caputo}:} Similar to the first downstream task, we compare our method to the multi-tasking approach JiGen.

\textbf{Registered:} Because the CT and MR modalities are not registered in this benchmark, we register the modalities in this baseline. This aims to assess the influence of registration on learned representations by our multimodal puzzles. We employ VoxelMorph~\cite{voxel_morph}\footnote{Our aim is to benchmark our method against a proven image registration method (VoxelMorph takes structural information into consideration)} for multimodal image registration.

\textbf{2D Models Genesis~\cite{ssl_models_genesis} (2D MG):} Similar to the first task, we fine-tune this model on multimodal Liver data. 

\textbf{2D Rubik Cube~\cite{rubik} (2D RC):} Similar to the first task, we fine-tune this method on multimodal liver data.

\textbf{U-Net Puzzle Solver:} This baseline aims to evaluate the representations learned by our puzzle solver, which relies on Sinkhorn operator, in comparison to a U-Net puzzle solver, which relies on pixel-wise image reconstruction.

\paragraph{Evaluation Metrics}
We report the results of liver segmentation using the average dice score (Dice). This metric is used on the official challenge. 

\paragraph{Discussion}
The results of our \textbf{multimodal} method compared to the above baselines are shown in table~\ref{tab_liver}. Our method outperforms both "from scratch" and "single-modal" baselines in this task too, supporting the advantages of initializing the model using our multimodal puzzle solving task. We also compare with 2D Models Genesis (2D MG)~\cite{ssl_models_genesis}, which we outperform in this downstream task, supporting the effectiveness of our pretraining method. However, we observe that our method only marginally outperforms this method, and we believe this is because Models Genesis was pretrained on another Chest CT data.
Also, our method outperforms the multitasking method JiGen~\cite{domain_adaptation_caputo}, when trained on this task too. The results against the "Registered" baseline are almost on par with the results of our "multimodal" method using non-registered data. This result is significant because it highlights our multimodal puzzles' ability to operate on non-registered imaging modalities. Similar to the first downstream task, our Sinkhorn-based puzzle solver outperforms the U-Net puzzle solver, confirming that the representations obtained by our solver hold more semantic information. Compared to 2D Rubik Cube~\cite{rubik}, we outperform this method on this downstream task too. This confirms that our method is able to learn better multimodal representations, given the same input modalities.
\begin{table}
\begin{center}
\centering         \small
\caption{Results on the Liver segmentation task}\label{tab_liver}
\begin{tabular}[t]{ c c c } \toprule
Model & Avg. Dice & \textit{p}-value \\
\midrule
From scratch & 89.98 & 1.2$e$-5\\
Registered & 95.09 & 9.6$e$-1 \\
Single-modal & 92.01 & 6.3$e$-3 \\
\hline 
JiGen~\cite{domain_adaptation_caputo} & 93.18 & 2.1$e$-2  \\
2D MG~\cite{ssl_models_genesis} & 95.01 & 3.8$e$-1 \\
2D RC~\cite{rubik} & 94.15 & 4.9$e$-2 \\
U-Net Solver & 85.38 &  5.9$e$-13 \\
\hline 
Ours (Multi-modal) & \textbf{95.10} & - \\
\bottomrule
\end{tabular}
\end{center}
\end{table}

%-------------------------------------------------------------------------
\subsubsection{Survival Days Prediction (Regression)} 
The BraTS challenge involves a second downstream task, which is the prediction of survival days. The number of training samples is 60 cases, and the validation set contains 28 cases. Similar to what we did for the other downstream tasks, we transfer the learned weights of our multimodal puzzle solver model. The downstream task performed here is regression, hence the output of our trained model here is a single scalar that represents the expected days of survival. We reuse the convolutional features, and we add a fully connected layer with five features in it, and then a single output layer on top. We also include the age as a feature for each subject right before the output layer. The size of the fully connected layer, was determined based on the obtained performance, i.e. by hyperparameter tuning.

In table~\ref{tab_reg}, we compare our results to the baselines of Suter~\emph{et al.}~\cite{suter}. In their work, they compared deep learning-based methods performances with multiple other classical machine learning methods on the task of survival prediction. The first experiment they report is \textbf{(CNN + age)}, which uses a 3D CNN. The second is a \textbf{random forest regressor}, the third is a multi-layer perceptron (MLP) network that uses a set of hand-crafted features called \textbf{FeatNet}, and finally, a \textbf{linear regression} model with a set of 16 engineered features. We outperform their results in all cases when fine-tuning our puzzle solver model on this task. The reported evaluation metric is the Mean Squared Error.
\begin{table}
\begin{center}
\centering           \small
\caption{BraTS survival prediction (regression). The baselines (except for "From scratch") are taken from~\cite{suter}}\label{tab_reg}
\begin{tabular}[t]{ c r } \toprule
Model & MSE \\
\midrule
From scratch & 112.841 \\ 
\hline
CNN + age & 137.912 \\
Random Forest Regression & 152.130 \\
FeatNet + all features & 103.878 \\ 
Lin. Reg. + top 16 features & 99.370 \\ 
\hline
Ours (Multi-modal) & \textbf{97.291} \\
\bottomrule
\end{tabular}
\end{center}
\end{table}

%------------------------------------------------------------------------
\subsection{Cross-Modal Generation Results} \label{cross_modal} 
We investigate in this set of experiments the effect of the cross-modal generation step. As mentioned earlier, obtaining large multimodal medical imaging datasets can be sometimes challenging. Therefore, we investigate in this set of experiments, the effect of the explicit cross-modal generation step. This step allows for better adoption of our multimodal puzzle-solving, even in the case of a few multimodal samples only. It is common that some imaging modalities exist in larger quantities than others. Hence, in this set of experiments, we perform this step in a semi-supervised fashion, assuming small multimodal and large single-modal data subsets. The size of the multimodal subset affects the downstream task performance, and the quality of generated data.
We evaluate the generation process at data subset sizes of 1\%, 10\%, and 50\% of the total number of patients in each benchmark. 
We assume a reference modality, which is often abundant in practice, to generate the other modalities. When there is no clear reference modality, it is also possible to generate all modalities from each other, which results in an increased number of trained GAN models. 
In the BraTS and Prostate benchmarks, we use T2-weighted MRI. In the Prostate dataset, we use T2-weighted MRI scans to generate the ADC diffusion-weighted scans. In BraTS, since we have four MRI modalities, we train three GANs and convert T2 MRI to the other MRI modalities (T1, T1CE, FLAIR). In the CHAOS liver benchmark, we use the CT modality to generate T2 MRI.

\begin{figure*}
\begin{center}
    \includegraphics[width=\linewidth]{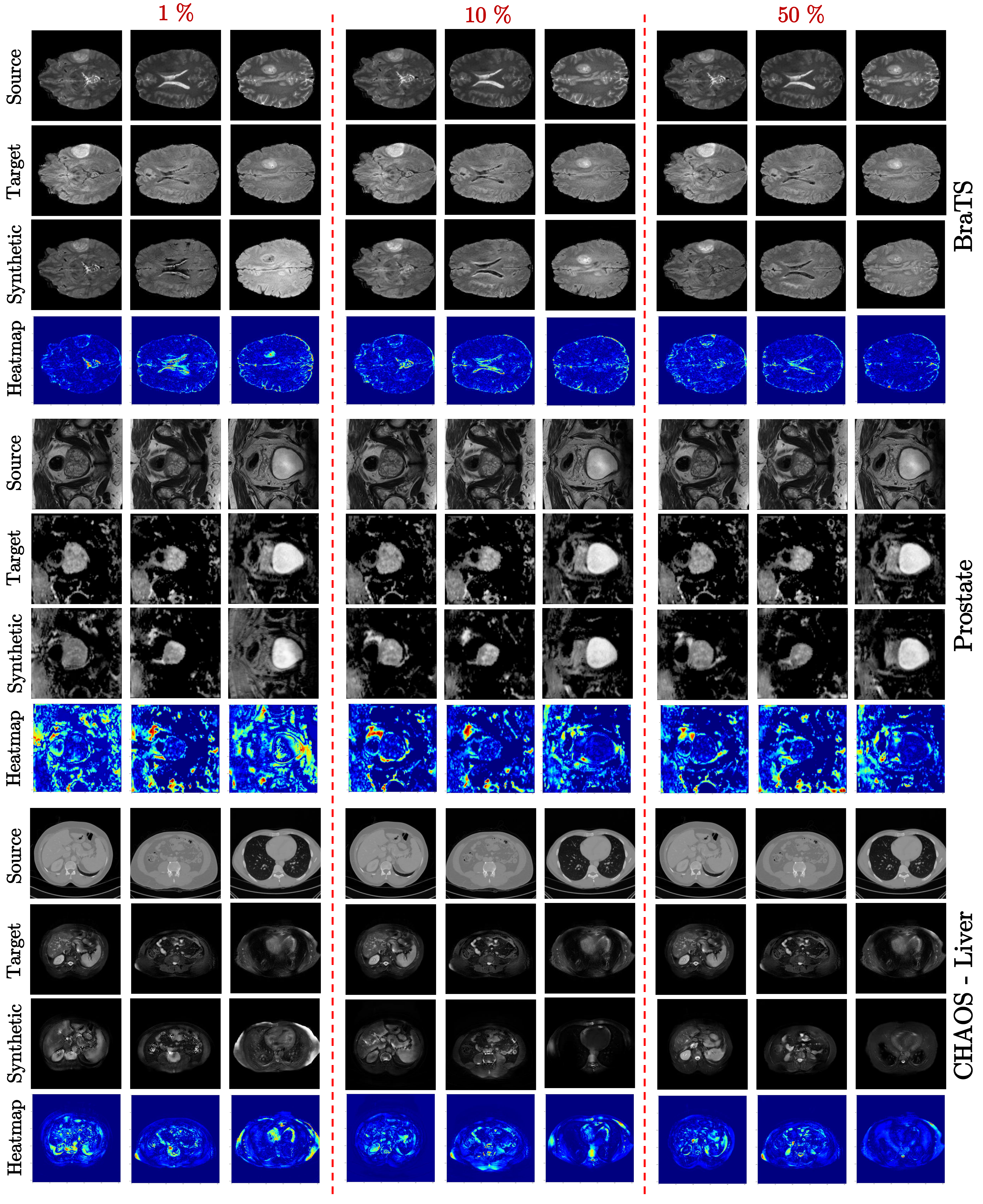}    
\end{center}
\caption{Qualitative results of the trained CycleGAN at different rates of multimodal data. The size of the multimodal set affects the quality of generated data, and the performance on downstream tasks. We evaluate the generation process at data subset sizes of 1\%, 10\%, and 50\% of the full data size in each benchmark. The figure shows converted scans from T2 to FLAIR for BraTS, T2 to ADC for Prostate, and CT to MR-T2 for CHAOS liver. Despite using small multimodal training set, the quality of synthetic images is high. The target images of the CHAOS liver dataset are obtained after applying the registration step with VoxelMorph.} \label{fig_ql}
\end{figure*}

\paragraph{Discussion.} This step is only justified if it provides a performance boost over the \textbf{single-modal} puzzle solving baseline, i.e. training our model on puzzles that originate from one modality. We measure the performance on the three downstream tasks, by fine-tuning these models and then evaluating them on segmentation. 
The presented results in table~\ref{tab_gan} clearly show an improvement on all benchmarks, when training our puzzle solver on the mixture of synthetic and realistic multimodal data. Even when we use only 1\% of the total dataset sizes, the generator appears to capture the important characteristics of the generated modality. The qualitative results in figure~\ref{fig_ql}, confirm the quality of generated images.
In addition, we study the benefits of using the synthetic data for self-supervised pretraining, instead of training the downstream task directly on them. The results of \textbf{Our method} in table~\ref{tab_gan} support this approach, as opposed to those of direct \textbf{Downstream training}.

\begin{table*}
\begin{center}
\centering        \small
\caption{Results on segmentation in avg. dice scores. The percentages are sizes of multimodal subsets used to train CycleGAN }\label{tab_gan}
\begin{tabular}[t]{ c c c c c c c } 
    \toprule
    \multirow{2}{*}{Model} & \multicolumn{3}{c}{BraTS} & \multicolumn{2}{c}{Prostate} & CHAOS\\
    \cline{2-7}
     & ET & WT & TC & C & P & Liver\\
    \midrule
    Single-modal & 72.12 & 82.72 & 79.61 & 69.48 & 87.42 & 92.01 \\
    \hline 
    Downstream training (1\%) & 65.40 & 74.11 & 69.24 & 55.24 & 71.23 & 80.31 \\
    Downstream training (10\%) & 69.28 & 78.72 & 71.58 & 62.65 & 76.18 & 83.65 \\
    Downstream training (50\%) & 72.92 & 81.20 & 78.36 & 66.34 & 80.24 & 87.58 \\
    \hline 
    Our method (1\%) & 73.12 & 82.42 & 80.01 & 61.87 & 82.67 & 82.71 \\
    Our method (10\%) & 74.19 & 85.71 & 81.33 & 67.67 & 84.37 & 86.26 \\
    Our method (50\%) & \textbf{76.23} & \textbf{87.04} & \textbf{82.38} & \textbf{73.45} & \textbf{87.92} & \textbf{93.85} \\
    \bottomrule
\end{tabular}
\end{center}
\end{table*}

%------------------------------------------------------------------------
\subsection{Low-Shot Learning Results} \label{low_shot} 
\begin{figure*}
\begin{center}
    \includegraphics[width=\linewidth]{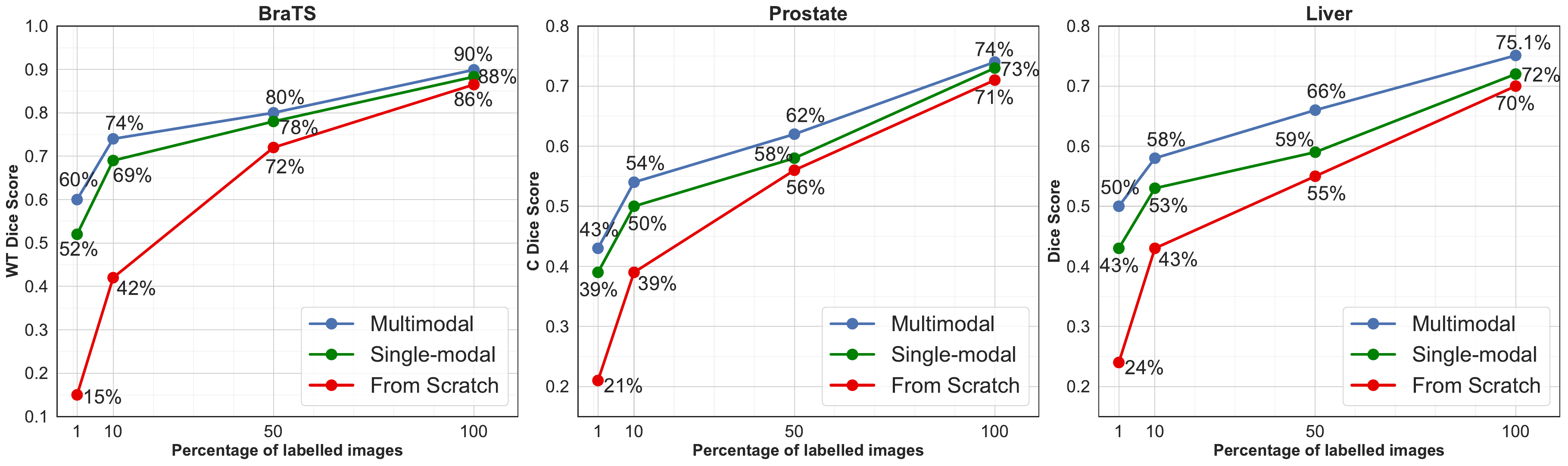}    
\end{center}
\caption{Results in the low-shot scenario. Our method outperforms both "single-modal" and "from scratch" baselines, confirming the benefits in data-efficiency when pretraining using our multimodal puzzles.} \label{plot_few_shot}
\end{figure*}

In this set of experiments, we assess how our self-supervised task benefits data-efficiency of the trained models, by measuring the performance on both downstream segmentation tasks at different labeling rates by fine-tuning our pre-trained model with corresponding sample sizes. We randomly select subsets of patients at 1\%, 10\%, 50\%, and 100\% of the total segmentation training set size. Then, we fine-tune our model on these subsets for a fixed number of epochs (50 epochs each).
Finally, for each subset, we compare the performance of our fine-tuned \textbf{multimodal} model to the baselines trained \textbf{from scratch} and \textbf{single-modal}. As shown in figure~\ref{plot_few_shot}, our method outperforms both baselines with a significant margin when using few training samples. This gap to the \textbf{single-modal} baseline confirms the benefits of using our multimodal puzzles instead of traditional single-modal puzzles.
In a low-data regime of as few samples as 1\% of the overall dataset size, the margin to the \textbf{from scratch} baseline appears larger. This case, in particular, suggests the potential for generic unsupervised features applicable to relevant medical imaging tasks. Such results have consequences on annotation efficiency, i.e. only a fraction of data annotations is required. It is worth noting that we report these low-shot results on \emph{realistic} multimodal data. The \textbf{single-modal} baseline uses these modalities for each task: FLAIR for BraTS, T2 for Prostate, and CT for Liver.

%------------------------------------------------------------------------
\subsection{Ablation Study} \label{ablation}
We use realistic data only in the experiments presented in this ablation study. 

\subsubsection{Puzzle Complexity} \label{puzzle_complexity}
\begin{figure}
\begin{center}
    \includegraphics[width=\linewidth]{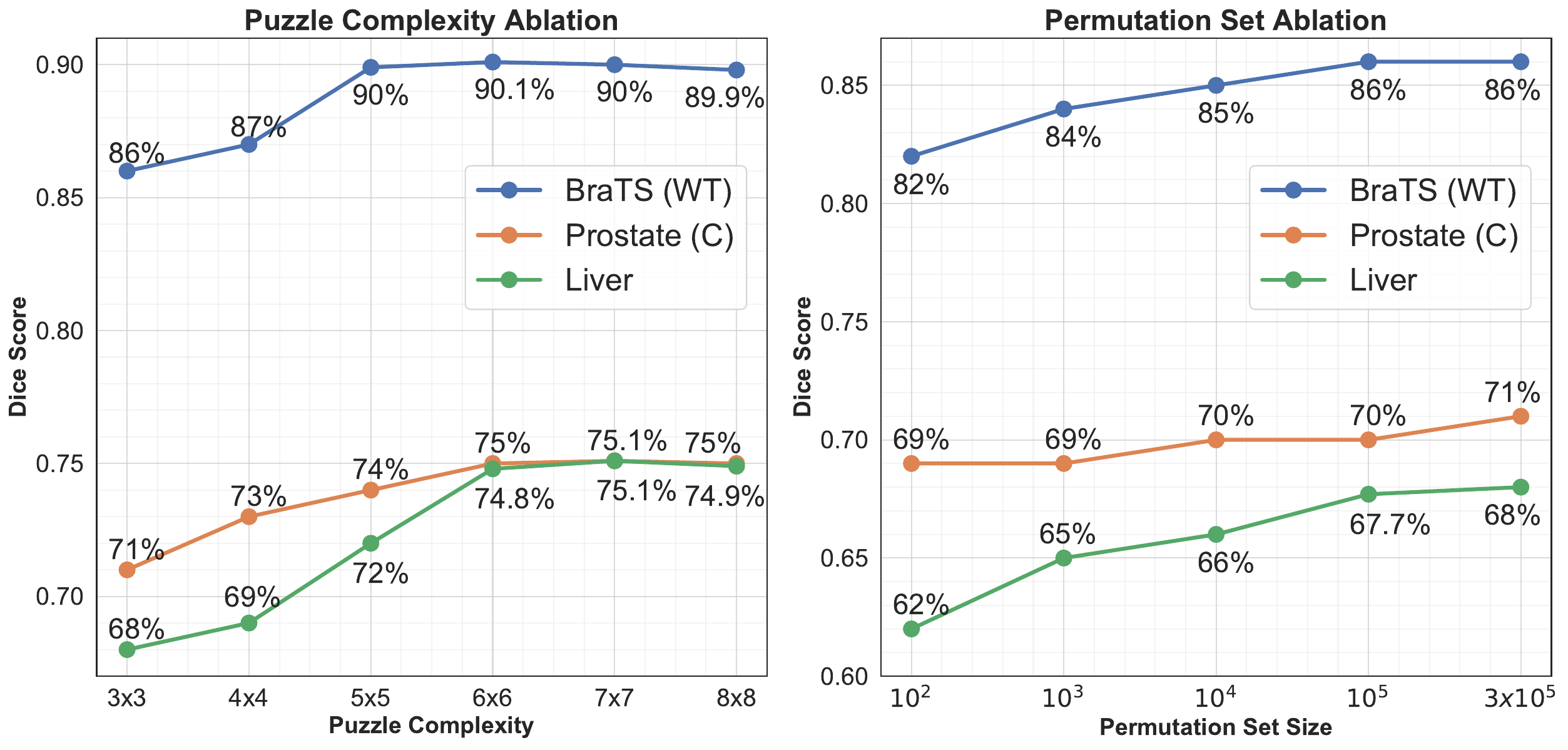}
\end{center}
\caption{(Left) Puzzle complexity vs. downstream performance. The trendlines suggest more complex jigsaw puzzles improve downstream task performance, up to a certain point where additional complexity has little effect. (Right) Permutation set size vs. downstream performance. The plot shows that a larger set than a certain size has little influence on downstream performance, but a small set affects the results negatively.} \label{plot_ablation}
\end{figure}

In this set of experiments, we analyze the impact of the complexity of our multimodal jigsaw puzzles in the pretraining stage, on the performance of downstream tasks. This aims to evaluate whether the added complexity in this task can result in more informative data representations; as the model is enforced to work harder to solve the more complex tasks. Our results confirm this intuition, as shown in figure~\ref{plot_ablation} (Left), where, in general, the downstream task performance (measured in Dice Score) increases as the puzzle complexity rises. This is correct up to a certain point, at which we observe that the downstream performance almost flattens. This complexity turning point is different across the three downstream tasks, as can be seen in figure~\ref{plot_ablation} (Left). We should note that in our previous experiments, we use $5\times5$ puzzle complexity for BraTS tasks, and $7\times7$ for Prostate and Liver.

\subsubsection{Permutation Set Size} \label{perm_list}
In this set of experiments, we analyze the impact of the permutation list size used in creating multimodal jigsaw puzzles, on the performance of downstream tasks. This hyperparameter can be viewed as another form of puzzle complexity~\cite{jig}. However, as mentioned before, our design choice to utilize the efficient Sinkhorn operator for puzzle solving captures the whole set of possible permutations. The main reason lies in solving the task as permutation matrix inference (regression), instead of permutation classification used in other solutions~\cite{jig,rubik,domain_adaptation_caputo}. Consequently, our puzzle solver is expected to reduce the influence of the permutation list size on trained models. Nevertheless, in practice, we sample from a finite permutation set, for computational reasons, and we attempt to employ a permutation set as large as possible. Our intuition here is that a larger permutation set can facilitate learning a better semantic representation. Hence, in this set of experiments, we analyze the effect of the chosen permutation set size. Figure~\ref{plot_ablation} (Right) shows the downstream performance for every permutation set size. It is worth noting that in this plot, we use a 3-by-3 puzzle complexity, which results with $9!$ total permutations. We vary the permutation set size from $10^2$ to $3\times10^5$, which is almost equal to $9!$. The trendlines in the plot show that a small permutation set, e.g. 100 or 1000, can harm the learned representations, as the model may overfit this small set of permutations. It also shows that an increase in permutation set size has a positive effect on downstream performance, similar to puzzle complexity. 
Nevertheless, the influence of permutation set size becomes negligible after a certain limit, as appears in the figure~\ref{plot_ablation}. This is consistent with the effect of varying puzzle complexity. 
We should highlight that the actual number of permutations grows with the number of modalities in each downstream task, as explained in Section~\ref{puzzle_construction}. However, since our method learns cross-modal representations, it allows for cancelling out this exponential growth with more modalities.

Across both ablations presented in this section (for puzzle complexity and permutation set size), we observe a consistent behavior, even though the actual numbers may differ. This behavior is that both hyperparameters improve downstream performance as they increase, up to a certain limit, where their variation causes a negligible change in performance then. We believe that the model capacity here is the cause of such saturation, which is consistent with the findings of Goyal \emph{et al.}~\cite{Scaling_self_supervision}. Employing a larger model architecture may benefit more from increased complexity in our multimodal puzzles. We deem evaluating several architectures as future work.

%-------------------------------------------------------------------------
\section{Conclusion \& Future Work} 
In this work, we proposed a \textit{multimodal} self-supervised Jigsaw puzzle-solving task. This approach allows for learning rich semantic representations that facilitate downstream task solving in the medical imaging context. In this regard, we showed competitive results to the state-of-the-art results in three medical imaging benchmarks. One of which has unregistered modalities, further supporting the effectiveness of our approach in producing rich data representations.
The proposed multimodal puzzles outperform their single-modal counterparts, confirming the advantages of including multiple modalities when constructing jigsaw puzzles.
It is also noteworthy that the efficient Sinkhorn operator enabled large permutation sets and puzzle complexities, an aspect commonly used puzzle solvers do not offer.
In addition, our approach further reduces the cost of manual annotation required for downstream tasks, and our results in the low-data regime support this benefit. 
We also evaluated a cross-modal translation method as part of our framework, which when used in conjunction with our method, it showed performance gains even when using few multimodal samples to train the generative model. Finally, we demonstrated the benefits of multimodality in our multimodal jigsaw puzzles, and we aim to generalize this idea to other self-supervised tasks. In addition, we believe generalizing our multimodal puzzles to the 3D context should improve the learned representations, and we deem this as future work.

\bibliographystyle{IEEEtran}
\bibliography{tmi}

\begin{appendices}
\section{Model Training for all tasks} \label{appendix_training_details}
\textbf{Input preprocessing.} For all input scans, we perform the following pre-processing steps:
\begin{itemize}
    \item First, we create 2-dimensional slices by navigating the scans from all datasets over the axial axis ($z$-axis).
    \item We resize each slice to a resolution of $128\times128$ for samples from BraTS, $256\times256$ for both Prostate and CHAOS.
    \item Then, each slice's intensity values are normalized by scaling them to the range $[0,1]$.
\end{itemize}

\textbf{Training details.} For all tasks, we use Adam~\cite{Adam_supp} optimizer to train our models. The initial learning rate we use is $0.001$ in puzzle solving tasks, $0.0002$ in cross-modal generation tasks, and $0.00001$ for segmentation and regression tasks. The network weights are initialized from a Gaussian distribution of $\mathcal{N}\left(\mu=0.1, \sigma=0.001\right)$ in puzzle solving and segmentation tasks, and from the distribution $\mathcal{N}(\mu=0, \sigma=0.02)$ in the cross-modal generation task. An $L_2$ regularizer with a regularization constant $\lambda=0.1$ is imposed on the network weights in puzzle solving and downstream tasks. In terms of training epochs, we train all the puzzle solving tasks for 500 epochs, the cross-modal generators for 200 epochs, and all fine-tuning on downstream tasks for 50 epochs.

\textbf{Network architectures.} All of our network architectures are convolutional, and they vary in small details per task:
\begin{itemize}
    \item For jigsaw puzzle solving tasks: we use 5 convolutional layers, followed by one fully-connected layer and one Sinkhorn layer. 
    \item For downstream segmentation tasks: we use a U-Net~\cite{UNET} based architecture, with 5 layers in the encoder, and 5 layers in the decoder. When fine-tuning, the weights of the encoder layers are copied from a pretrained model. The decoder layers, on the other hand, are randomly initialized. In terms of training losses in these tasks, we utilize a combination of two losses: i) weighted cross-entropy, ii) dice loss. We use the same importance to both losses in the total loss formula.
    \item For cross-modal generation tasks: as mentioned earlier, we largely follow the architecture of the \texttt{CycleGAN}~\cite{cyclegan} model. For the \emph{generators}, we use the Johnson \emph{et al.}'s~\cite{johnson_supp} architecture. We use 6 residual blocks for $128 \times 128$ training images, and 9 residual blocks for $256 \times 256$ or higher-resolution training images. With regards to the network \emph{discriminators}, we utilize the PatchGAN~\cite{PatchGAN_supp} discriminator architecture, which processes $70\times70$ input patches.
\end{itemize}

\textbf{Processing multi-modal inputs.} 
In Brain tumor and Prostate segmentation tasks, the reported methods from literature use all available modalities when performing the segmentation, e.g. in table~\ref{tab_brain} in our paper. They typically stack these modalities in the form of image color channels, similar to RGB channels. However, our proposed puzzle-solving method expects a single channel input at test time, i.e. one slice with multi-modal patches. This difference only affects the input layer of the pretrained network, as fine-tuning on an incompatible number of input channels causes this process of fine-tuning to fail. We resolve this issue by duplicating (copying) the weights of \emph{only} the pretrained input layer. This minor modification only adds a few additional parameters in the input layer of the fine-tuned model, but allows us to leverage its weights. The other alternative for this solution is to discard the weights of this input layer, and initialize the rest of the model layers from pretrained models normally. However, our solution for this issue takes advantage of any useful information encoded in these weights, allowing the model to fuse data from all the channels. The exact numbers of channels in each downstream task is as follows:
\begin{itemize}
    \item BraTS Brain Tumor Segmentation: in each input slice, the MRI 4 modalities are stacked as channels.
    \item BraTS Number of Survival Days Prediction: for each input slice we also stack the 4 MRI modalities, on top of the predicted tumor segmentation mask; summing up to 5 channels for each input slice. The predicted masks are produced by our best segmentation model.
    \item Prostate segmentation: we stack the 2 available MRI modalities in each input slice.
\end{itemize}
In the Liver segmentation task, however, stacking the input modalities as channels is not possible. This is due to the fact that the modalities in this task (CT and MR-T2) are non-registered. Hence, we process these modalities using the Joint Learning scheme used in~\cite{multimodal_unpaired}. This means we process each modality as a single input slice. Pretrained models using our multimodal puzzles, 
learns to disregard the modality type during training and testing.

\textbf{Training the multimodal puzzle solver}
It is noteworthy that after we sample patches from input slices, we add a random jitter of 5 pixels in each side before using them in constructing puzzles. This mechanism ensures the model does not use any shortcuts in solving the puzzles, thus enforcing it to work harder and learn better representations.

Algorithm~\ref{algo2} provides the detailed steps of the training process of our proposed multimodal puzzle solver. After obtaining the network parameters, the yielded representations capture different tissue structures across the given modalities as a consequence of the multimodal puzzle solving. Therefore, they can be employed in downstream tasks by simply fine-tuning them on target domains. 

\begin{algorithm} 
    \SetKwProg{Algorithm}{Algorithm}{}{}
    \SetKwProg{Procedure}{Procedure}{}{}
    \SetNlSty{textbf}{}{:}
    \DontPrintSemicolon
    
    \Algorithm{\textsc{Train Puzzle Solver}}{
      \KwIn{list of multimodal $puzzles$}
      \KwOut{trained model $G$}
      $G \leftarrow$ initialize model weights $w$\;

        \ForEach(\tcp*[h]{each puzzle contains $N$ patches}){$P$ from $puzzles$}{
          \ForEach{patch $x$ in $P$}{
            $v \leftarrow G(x)$ \tcp*[h]{$N$-dimensional feature vector}
          }
          $V \leftarrow$ concat. vectors $v$ \tcp*[l]{form a matrix with size $N \times N$} 
          $S \leftarrow Sinkhorn(V)$ \tcp*[l]{permutation matrix} 
          $P_{rec} \leftarrow S^T.P$ \tcp*[l]{reconstructed version} 
          $loss \leftarrow MSE(P^*,P_{rec})$
        }
    } 
    \caption{One epoch of training multimodal puzzle solver} \label{algo2}
\end{algorithm}

\end{appendices}

\end{document}